%% file: main.tex
\documentclass[runningheads]{llncs}
\usepackage{amsmath}
\usepackage{amssymb}
\usepackage{multirow}
\def\maketag@@@#1{\hbox{\m@th\normalfont\normalsize#1}}
\newcommand\Label[1]{&\refstepcounter{equation}(\theequation)\ltx@label{#1}&}
\makeatother
\usepackage{tikz}
\usetikzlibrary{fit,positioning,arrows,automata,calc}
\tikzset{
	main/.style={circle, minimum size = 5mm, thick, draw =black!80, node distance = 10mm},
	connect/.style={-latex, thick},
	box/.style={rectangle, draw=black!100}
}
\usepackage{enumitem}
\usepackage{caption}
\usepackage{subcaption}
\usepackage{scalerel}
\begin{document}
\title{Variational Conditional Dependence Hidden Markov
	Models for Skeleton-Based Action Recognition\thanks{This work has received funding from the European Union’s Horizon 2020 research and innovation program under grant agreement No 872139, project aiD. }}
\titlerunning{Variational Conditional Dependence HMM}
%
\author{Konstantinos P. Panousis\inst{1} \and
Soritios Chatzis\inst{1} \and
Sergios Theodoridis\inst{2,3}}
%
\authorrunning{K.P. Panousis et al.}
%
\institute{Cyprus University of Technology, Limassol Cyprus\\ \email{k.panousis@cut.ac.cy}\\ \and
National and Kapodistrian University of Athens, Athens, Greece\and
Aalborg University, Denmark}
\maketitle              

\begin{abstract}
	Hidden Markov Models (HMMs) comprise a powerful generative approach for modeling sequential data and time-series in general. However, the commonly employed assumption of the dependence of the current time frame to a single or multiple immediately preceding frames is unrealistic; more complicated dynamics potentially exist in real world scenarios. This paper revisits conventional sequential modeling approaches, aiming to address the problem of capturing time-varying temporal dependency patterns. To this end, we propose a different formulation of HMMs, whereby the dependence on past frames is dynamically inferred from the data. Specifically, we introduce a hierarchical extension by postulating an additional latent variable layer; therein, the (time-varying) temporal dependence patterns are treated as latent variables over which inference is performed. We leverage solid arguments from the Variational Bayes framework and derive a tractable inference algorithm based on the forward-backward algorithm. As we experimentally show, our approach can model highly complex sequential data and can effectively handle data with missing values. 
	
\end{abstract}
\makeatother

\section{Introduction}

Modeling sequential data, typically encountered in many real-world applications such as bioinformatics and computer vision, remains a fundamental task in the field of machine learning.
Hidden Markov Models have been a popular approach for modeling such data, but more recently, have largely been replaced by their ``deep'' variants. Despite their success, these approaches exhibit significant drawbacks such as over-parameterization even in simple tasks as image classification\cite{panousis}. Nevertheless, both models aim to learn hidden representations for sequential data.

Commonly, first-order HMMs are usually considered; this allows for simplicity and low computational complexity. Then, model temporal dynamics are restrained to a simple step back, undermining model effectiveness. This compromise is especially detrimental in many applications where longer temporal dynamics may be present. Researchers have considered alternatives to alleviate this restriction, for example by introducing extended temporal dependencies in the form of second or higher order Markovian dynamics and had successful applications in many domains \cite{mariRobotics,Nel,ENGELBRECHT}. 
However, the increased order introduces three significant drawbacks: (i)  additional complexity to the model that may render it unusable in real-world scenarios; (ii) unnecessary burden to the researchers, namely the need to determine the best postulated order for each specific application and dataset; and (iii) unnecessary complex models with strong overfitting tendencies, which may negate the benefits and flexibility of HMMs. 

Lastly, a further drawback of conventional HMM approaches is the static and homogeneity assumptions \cite{Begleiter,chatzisInfinitely,chatzisNonstationary}, where the possible dynamic temporal dependencies of the data are ignored. The same effect applies to hidden semi-Markov models (HSMMs) \cite{hsmm}; even though the relaxed temporal assumptions  allow for more flexible modeling of the dynamics, potential non-homogeneous temporal dynamics in the data are still ignored.  Drawing inspiration from there results, researchers have developed HMMs with variable order Markov chains, e.g., \cite{buhlmann1999}. The resulting models have been shown to be effective in a diverse range of applications, nevertheless exhibiting significant drawbacks such as the inability to model continuous observations \cite{dimitrakakis10a}.

Recently, \cite{hdm} introduced a variant of a simple HSMM model, dubbed Hierarchical Dynamic Model (HDM). Specifically, the proposed model constitutes a hierarchical extension by leveraging on the Bayesian framework to increase the capacity of HSMMs to account for temporal and spatial variations in the Human Action Recognition task. The produced generative model is shown to be more robust to the natural variations of the data; it exhibits increased generalization capabilities, while at the same time requiring less data to train.

This paper draws from these results and attempts to offer a principled way of modeling sequential data with complex dynamics. To this end, we develop a novel variant of HMMs that is able to capture complex temporal dependencies.  The considered approach constitutes a hierarchical model: by postulating an \textit{additional latent first-order Markov Chain}, called the \textit{dependence-generator} layer, the model can alter the effective temporal dynamics of the conventional \textit{observation-emitting} Markov Chain. In this way, the proposed approach can effectively infer which past state more strongly affects the current time frame. The proposed inferential construction is enhanced by a fully Bayesian treatment under the Variational Inference framework. We posit that the proposed hierarchical variant, combined with the flexibility and effectiveness of \textit{Variational Bayes}, can greatly increase the capacity of the resulting architecture to model sequential data that exhibit a complex combination of spatial and temporal variations.

We formulate efficient training and inference algorithms for our approach by: (i) deriving a variant of the classical forward-backward algorithm used in HMMs, (ii) relying on Variational Inference and conjugate priors for closed-form solutions. We dub our approach \textit{Variational Bayesian Conditional Dependence Hidden Markov Model} (VB-CD-HMM) and evaluate its performance in the Human Action Recognition domain. Human Action Recognition poses one of the most challenging tasks in the computer vision community.  Researchers have devoted significant effort to address this particular area \cite{surveyAction} and remarkable progress has been made with the popularization of action recognition through 3D data \cite{review3d}. In this context, various approaches have been employed in order to model the activity recognition task, such as HMM-based approaches \cite{hdm}, LSTM-based \cite{zhao_iccv} and Gaussian Processes \cite{chatzis_gaussian,gplvm}. However, the significant spatial and temporal variations arising in the execution of an action, along with variations due to the capturing process, e.g., camera occlusion, leave room for needed improvements in existing approaches. We provided strong empirical evidence for our approach though a series of experiments dealing with Human Action Recognition benchmarks. Our evidence shows competitive predictive accuracy, persistent even in the presence of missing values.
\section{Proposed Model}
In this work, we consider an extension of a conventional first-order HMM, whereby a hierarchical structure is employed consisting of \textit{two layers}: (i) a \textit{dependence-generator} layer; this comprises a simple first-order Markov Chain that determines the \textit{steps-back} taken into account in the second layer; and (ii) a chain of \textit{observation-emitting} hidden states, where the temporal dependencies are determined from the output of the \textit{dependence-generator} layer. The \textit{temporal dependencies} are \textit{inferred} in a data-driven fashion. An outline of the proposed Conditional Dependence HMM (CD-HMM) model is provided in Fig. \ref{graph:cdhmm}.

\begin{figure}[h]
	\centering
	\begin{subfigure}[b]{0.35\textwidth}
		\centering
		\resizebox{1.\linewidth}{!}{\input{graphs/cdhmm_graph}}
		\caption{A graphical illustration of the Conditional Dependence Hidden Markov Model.}
		\label{graph:cdhmm}
	\end{subfigure}     \hfill
	\begin{subfigure}[b]{0.6\textwidth}
		\centering
		\resizebox{1.\linewidth}{!}{\input{graphs/vb_cdhmm}}
		\caption{The resulting hierarchical VB-CD-HMM.}
		\label{graph:vbcdhmm}
	\end{subfigure}
\end{figure}

\subsection{Model Definition}
Let $Y = \{ \boldsymbol y_t \}_{t=1}^T \in \mathbb{R}^D$ be an observed data sequence with $T$ time frames and $D$ features. Following the definition of conventional HMMs, let us assume an $N$-state \textit{emitting state sequence} denoted as $X = \{x_t\}_{t=1}^T, x_t \in [1,\dots, N]$ where $x_t$ indicates the state from which the $t^{th}$ observation is emitted. Each emission density is modeled by an $M$-component finite  mixture model; the \textit{mixture component indicators} are denoted as $L = \{ l_t \}_{t=1}^T , l_t \in [1, \dots, M]$ with $l_t$ indicating which component generated the $t^{th}$ observation. 

The hierarchical construction postulates an additional layer, the \textit{dependence-generator} layer, such that the \textit{latent data} associated with each sequence are augmented with the supplementary sequence of \textit{temporal dependence indicators}, $Z = \{ z_t \}_{t=1}^T , z_t \in [1,\dots, K]$. These denote the current temporal dependencies between the \textit{observation-emitting} states at time $t$ and times $t-1, \dots, t-K$ as shown in Fig. \ref{graph:cdhmm}. Thus, different pairwise state dependencies $(x_t, x_{t-z_t})$ are valid at each time point $t$, dictated by the inferred latent variables $z_t$.

The model parameters comprise $\theta = \{\phi, \psi\}$, where $\phi$ contains the parameters of the emission distributions of the model and $\psi$ the effective parameters of the latent processes.

For the emission distributions, finite mixture models of multivariate Gaussian distributions are considered,
$
p(\boldsymbol y_t | x_t = i) = \sum_{m=1}^M c_{im} \mathcal{N}(\boldsymbol y_t| \boldsymbol\mu_{im},  R_{im}), \forall i 
$
where $\mathcal{N}(\cdot|\boldsymbol\mu,  R)$ is a multivariate Normal distribution with mean $\boldsymbol\mu \in \mathbb{R}^{D}$ and precision matrix $R \in \mathbb{R}^{D\times D}$ with distinct parameters for each state and mixture; $\boldsymbol{c}_{i} \in \mathbb{R}^{M}$ are the mixture component coefficients for state $x_t =i$.  Hence, the effective parameters for the model distributions are $\phi = \{ \boldsymbol\mu_{im},  R_{im}, c_{im}\}_{i, m = 1}^{N, M}$. 

We proceed with the definition of the parameters and their relationships pertaining to the postulated latent processes. Specifically, the first, \textit{dependence-generator}, layer is a first-order Markov Chain and is described by the following initial and transition probabilities:
\begin{align}
	\hat{\pi}_k &\triangleq p(z_1 = k), \quad \forall k \\
	\hat{A}_{kk'} &\triangleq p(z_t = k' | z_{t-1} = k), \quad \forall t>1,k,k'
\end{align}
For the second, \textit{observation-emitting}, layer process, the initial probability of the emitting state reads:
\begin{align}
	\pi_i \triangleq p(x_1 = i), \quad \forall i
\end{align}
To model the dependency between the first and second layers, we define the set of \textit{conditional dependence} transition probability matrices $\{ A^k\}_{k=1}^K$ on the second-layer process, where $A^k \triangleq [A_{ij}^k]_{i,j=1}^N$,
\begin{align}
	\begin{split}
		A_{ij}^k &\triangleq  p(x_t = j | x_{t-k} = i, z_t = k) 
	\end{split}
\end{align}
Thus, based on the inferred dependence of the \textit{dependence-generator} layer, we consider different temporal dependencies in the second layer at each time point; these are regulated through the $z_t$ variables, which are imposed a first-order Markov chain themselves. Intuitively, the conditional dependence mechanism determines the strength of the dependence of the current state to individual \textit{past states}. The strongest inferred dependence is retained from the model, and is encoded in the $z_t$ values; thus, the model dynamically alters the transition dynamics for the  \textit{observation-emitting} state at each time point.

Having fully defined the parameters for the latent processes, the corresponding set comprises $\psi = \{ \hat{\boldsymbol\pi}, \hat{A}, \boldsymbol\pi, \{A^k\}_{k=1}^K \}$. The definition of the model is now concluded. 

The joint distribution reads:
\begin{align*}
	\begin{split}
		&p(Y, X, Z) =  \hat{\pi}_{z_1} \pi_{x_1} \prod_{t=1}^{T-1} \hat{A}_{z_t, z_{t+1}} \prod_{t>1} A_{x_{t - z_t}, x_t}^{z_t}  \prod_{t=1}^T p(\boldsymbol y_t|x_t)
	\end{split}
\end{align*}
\subsection{Model Training}

We train the CD-HMM model presented in the previous section considering a \textit{variational Bayesian treatment}. The \textit{Variational Bayes} (VB) approach was chosen over MCMC inference methods, e.g., Gibbs Sampling, considering the negligible performance differences and the significantly lower computational complexity.  Sampling-based approaches encompass large computational complexity; this restricts the applicability of the model to either more complex applications and datasets or different settings (e.g., streaming).

The VB treatment of the model comprises the introduction of appropriate prior distributions over all model parameters and maximization of the resulting Evidence Lower Bound (ELBO) expression. To obtain closed-form solutions for the updates of the parameters, we employ conjugate priors; a choice accompanied by the benefits of lower computational complexity and interpretability \cite{theodoridis}.
\\[1em]
\noindent \textbf{Priors and Evidence Lower Bound.} 
Dirichlet distributions are imposed as the priors of the initial state probabilities for both layer processes $p(\boldsymbol\pi) = \mathcal{D}(\boldsymbol\pi | \boldsymbol\eta_0)$ and $p(\hat{\boldsymbol\pi}) = \mathcal{D}(\hat{\boldsymbol\pi} | \boldsymbol\alpha_0)$. Analogous Dirichlet priors are imposed on the rows of the state transition probabilities and the mixture coefficients of the emission distributions, i.e.,
{\small
	\begin{align}
		p(\boldsymbol{c}_{i,:}) &= \mathcal{D}(\boldsymbol{c_{i,:}}|\boldsymbol{w}_i), \ \boldsymbol{w}_i \in \mathbb{R}^M, \ i=1,\dots, N\\
		p(\hat{\boldsymbol{A}}_{k,:}) &= \mathcal{D}(\hat{\boldsymbol{A}}_{k,:}|\boldsymbol{\alpha}_k), \boldsymbol{\alpha}_k \in \mathbb{R}^K, k=1, \dots, K\\
		p(\boldsymbol{A}_{i,:}^k) &= \mathcal{D}(\boldsymbol{A}_{i,:}^k|\boldsymbol{\eta}_i^k), \boldsymbol{\eta}_i^k \in \mathbb{R}^N, i =1, \dots,N, \forall k
	\end{align}
}
For the prior over the means and precision matrices of the mixture components, we assign a Normal-Wishart distribution with hyperparameters $\lambda_{ij}, \boldsymbol m_{ij}, \eta_{ij},  S_{ij}$,
{\small
	\begin{align}
		\label{prior:nw}
		p( \boldsymbol\mu_{ij},  R_{ij}) = \mathcal{NW}(\boldsymbol \mu_{ij},  R_{ij}| \lambda_{ij}, \boldsymbol m_{ij}, \eta_{ij},  S_{ij}), \ \forall i,j
	\end{align}
}
This concludes the formulation of the prior specification for the VB treatment of the CD-HMM model, henceforth referred to as VB-CD-HMM. A graphical illustration of the fully proposed model can be found in Fig. \ref{graph:vbcdhmm}.

Let $\theta = \{\phi, \psi \}$ be the sets of latent process and emission distribution parameters where a conjugate exponential prior has been imposed. Under the VB treatment, we introduce an arbitrary distribution $q(\theta)$ and derive the Evidence Lower Bound (ELBO) for the model.
Maximizing the ELBO is equivalent to the minimization of the KL divergence between the true and the postulated posterior. We introduce the \textit{mean-field} (posterior-independence) assumption  on the joint variational posterior, such that $q(\theta)$ factorizes over all latent variables and model parameters. Then, the ELBO can be written as
{\footnotesize
	\begin{align}
		\label{prior:elbo}
		\begin{split}
			\text{ELBO} &= 
			 \mathbb{E}_{q(\theta), X, Y, Z} \Bigg[ \log \left(\hat{\pi}_{z_1}  \pi_{x_1} \prod_{t=1}^{T-1} \hat{A}_{z_t,z_{t+1}} \prod_{t>1}^{T}  A_{x_t,x_{t-z_t}}^{z_t}  \prod_{t=1}^T  p(\boldsymbol{y}_t|x_t) \right)\Bigg]\\
			&\quad + \mathbb{E}_{q(\theta)}\left[\log p(C, \pi, A, \hat{\pi}, \hat{A}, \boldsymbol\mu, \boldsymbol R) \right] - \mathbb{E}_{q(\theta)}\left[\log q(\theta)\right] \\
		\end{split}
	\end{align}
}
%
%
%
\\
\noindent \textbf{Variational Posteriors.} 
For each variable, the optimal member of the exponential family{\protect\footnote{By imposing conjugate priors, we obtain posteriors of the same functional form as their corresponding priors.}} can be found by maximizing \eqref{prior:elbo}. This maximization is performed in an EM-like fashion; in the E-step the parameters of the posterior distributions of the latent processes are updated, while in the M-step the rest of the variational distributions; the resulting iterative procedure is guaranteed to monotonically increase the ELBO. 
%
%
\\[1em]
\noindent \textbf{M-step.}
Starting from Eq. \eqref{prior:elbo}, we collect all relevant terms pertaining to the transition matrix of the first layer process $\hat{A}$; we then maximize the resulting expression, yielding the following variational posterior:
\begin{align}
	\label{eqn:bayesian_cdhmm:posterior_hat_pi}
	q(\hat{A}) = \prod_{i=1}^K \mathrm{Dir}(\hat{A}_{i1}, \dots, \hat{A}_{iK}|\omega_{i1}^{\hat{A}},\dots, \omega_{iK}^{\hat{A}})
\end{align}
where $\omega_{ij}^{\hat{A}} = \alpha_{ij} + \sum_{t} \gamma_{ijt}^{\hat{A}}$, $\gamma_{ijt}^{\hat{A}} \triangleq q(z_t=i, z_{t-1}=j)$. 

We follow the same procedure for the initial state probabilities, $\hat{\pi}$, the corresponding posteriors for the second layer process parameters $\pi$ and $A =\{A^k\}_{k=1}^K$, and the mixture component weights $C$. Due to conjugate priors, the emission distributions yield Normal-Wishart distributions with variational parameters $\{\tilde{\lambda}_{ij}, \tilde{\boldsymbol m}_{ij}, \tilde{\eta}_{ij}, \tilde{\boldsymbol S}_{ij}\}_{i,j=1}^{N,M}$; these are similarly obtained via standard computations, i.e., maximizing the corresponding ELBO expression.
\\[1em]
\noindent \textbf{E-step.} Turning to the E-step of the iterative procedure, the joint variational posterior optimizer for the latent processes $Z$ and $X$, and the mixture component indicators $L$ is analogous to the expression of the conditional probability defined in the context of conventional HMMs, e.g., see \cite{studentshmm}. Thus, the variational posterior entails the calculation of the corresponding responsibilities that comprise said posterior $q(z_1 = i)$, $q(x_1 =1)$, $q(z_t = i, z_{t-1} = j)$ and $q(x_t = j, x_{t-k} =i)$; these can easily be computed by means of of the well-known forward-backward algorithm. Thus, in the following, we derive a variant of the forward-backward algorithm for our model.
\\[1em]
\noindent \textbf{Forward-Backward Algorithm.}
\label{subsubsection_variant}
For calculating the forward-backward algorithm, we first define the forward messages as:
%
\begin{align}
	\alpha_t (\{x_\tau\}_{\tau = t-K+1}^t, z_t) \triangleq p(\{\boldsymbol y_\tau \}_{\tau=1}^t, \{x_\tau\}_{\tau =t-K+1}^t, z_t)
\end{align}
%
The defined messages can be computed recursively, using the following initialization:
{\normalsize
	\begin{align}
		\alpha_1(x_1, z_1) = 
		\begin{cases}
			\pi_i^* p^*(\boldsymbol y_1| x_1 = i), & z_1 =1 \\
			0, & z_1>1
		\end{cases}
	\end{align}
}
and the induction step reads:
{\small
	\begin{align}
		\begin{split}
			\alpha_t(\{x_\tau\}_{\tau =t-K+1}^t,  z_t) &= p^*(\boldsymbol y_t|x_t) \sum_{x_{t-K}}\sum_{z_{t-1}} \hat{A}_{z_{t-1}, z_{t}}^* A_{x_{t-k}, x_{t}}^{*z_t}
			\alpha_{t-1}(\{x_\tau\}_{\tau=t-K}^{t-1}, z_{t-1})
		\end{split}\label{eqn1}
	\end{align}
}
The backward messages are analogously defined as:
{\small
	\begin{align}
		\beta_t (\{x_{\tau}\}_{\tau = t-K+1}^{t}, z_t) \triangleq p(\{ \boldsymbol y_\tau\}_{\tau = t+1}^T | \{ x_\tau \}_{\tau = t-K+1}^t, z_t)
	\end{align}
}
The corresponding initialization reads:
\begin{align}
	\beta_T(\{x_\tau\}_{\tau=T-K+1}^T, z_T = k) = 1, \quad \forall k 
\end{align}
while the recursion becomes:
{\small
	\begin{align}
		\begin{split}
			\beta_t ( \{x_\tau \}_{\tau = t-K+1}^T, z_t) &= \sum_{x_{t+1}}\sum_{z_{t+1}} \hat{A}_{z_{t}, z_{t+1}}^* p^*(\boldsymbol y_{t+1}|x_{t+1}) \times \\
			&  A_{x_{t-k+1}, x_{t+1}}^{*z_{t+1}}
			\beta_{t+1}(\{x_\tau \}_{\tau = t-K+2}^T, z_{t+1})
		\end{split}
	\end{align}
}%
%
\\
\noindent \textbf{Responsibilities.}
By utilizing the forward-backward messages, the necessary responsibilities for updating the parameters of the model can now be computed.

For the first layer process, the marginal initial state responsibilities yield:
\begin{align*}
	\begin{split}
		\gamma_t^z(k)  &\triangleq p(z_t = k | \{\boldsymbol y_\tau\}_{\tau=1}^T) 
		\propto \sum_{X} \alpha_t(\{x_\tau\}_{\tau = t-K+1}^t, z_t) \beta_t(\{x_\tau\}_{\tau=t-K+1}^t, z_t)
	\end{split}
\end{align*}%
Analogously, for the \textit{observation-emitting} states, we have:
\begin{align*}
	\begin{split}
		\gamma_t^x(i) &\triangleq p(x_t = i | \{\boldsymbol y_\tau\}_{\tau=1}^T) \propto \sum_{X',z_t} \alpha_t(\{x_\tau\}_{\tau = t-K+1}^t, z_t) \beta_t(\{x_\tau\}_{\tau=t-K+1}^t, z_t)
	\end{split}
\end{align*}%
where $X' = \{x_\tau\}_{\tau = t-K +1}^t \setminus \{x_t\} =\{x_\tau\}_{\tau=t-K+1}^{t-1}$.

The state transitions responsibilities for the \textit{temporal dependence} indicators:
\begin{align*}
	\begin{split}
		\gamma_t^z (k,k') &\triangleq p(z_t = k', z_{t-1} = k|\{\boldsymbol y_\tau\}_{\tau=1}^T) \\
		&\propto \sum_{X'}\alpha_{t-1}(\{x_\tau\}_{\tau = t-K}^{t-1}, z_{t-1}) \hat{A}^*_{z_{t-1}, z_{t}}
		 \ A_{x_{t-k}, x_{t}}^{*z_t}p^*(\boldsymbol{y}_t | x_t) \beta_t (\{x_\tau\}_{\tau=t-K}^t, z_t)
	\end{split}
\end{align*}%
Finally, the dependent state transition responsibilities:
\begin{align*}
	\begin{split}
		\gamma_t^x (i,&j,k) \triangleq p(x_t=j, x_{t-k}=i| \{\boldsymbol y_\tau\}_{\tau=1}^T)
		\propto \sum_{X'', z_t}\alpha_t(\{x_\tau\}_{\tau=t-K+1}^t, z_t) \beta_t(\{x_\tau\}_{\tau=t-K+1}^t, z_t) 
	\end{split}
\end{align*}%
where $X'' = \{x_{\tau}\}_{\tau = t-K+1}^t \setminus \{x_t, x_{t-k}\}$.
\subsection{Inference}
Let us consider a VB-CD-HMM model, fully trained with data $Y$. We seek to calculate the predictive density of the test sequence $Y^{\mathrm{test}}$:
	\begin{align}
		\label{inference:predictive}
		p(Y^{\text{test}}|Y) = \int d\hat{\theta} \ p(\hat{\theta}|Y) P(Y^{\text{test}}|\hat{\theta})
	\end{align}
	Using the introduced variational posterior in place of the unknown true posterior $p(\hat{\theta}|Y)$ and similar to \cite{studentshmm}, we can compute the necessary quantities by utilizing the forward algorithm variant defined in Section \ref{subsubsection_variant}. Specifically, the density of a given test set $Y^{\text{test}} = \{\boldsymbol{y}_t^{\text{test}}\}_{t=1}^T$ yields:
	\begin{align}
		\label{eqn:predictive}
		p(Y^{\text{test}}|\hat{\theta}) = \sum_{z_T}\sum_{X} \alpha_T(x_{T-K+1}, \dots, x_T, z_T) 
	\end{align} 
	%
	%
	%
	\section{Experimental Evaluation}
	
	\subsection{Experimental Details}
	We use four popular benchmark datasets for the human action recognition task, namely \textit{MSR Action 3D (MSRA)} \cite{msra}, \textit{UTD-MHAD (UTD)} \cite{UTD}, \textit{Gaming 3D (G3D)} \cite{g3d} and \textit{UPenn Action (Penn)} \cite{Penn}. Only skeletal data are used for training the models. The preprocessing and augmentation of the data are the same as in \cite{hdm}. We extract the motion of the joints for every pair of joints between two consecutive frames. PCA is then employed for dimensionality reduction. 
	
	In addition, we further consider evaluation with a more complex dataset, namely NTU \cite{ntu}. This comprises $\approx 57000$ samples of $60$ different actions. We adopt the two data splits suggested in the original publication: (i) Cross-View and (ii) Cross-Subject.

	\subsection{Parameter Initialization \& Hyperparameter Selection}
	The resulting ELBO expression is non-convex with respect to the variational posterior, and potentially many local maxima exist; the obtained solution is thus dependent on parameter initialization. To avoid the rather time-consuming procedure of multiple runs from different initial values, we employ a common initialization scheme: K-Means is used to initialize the posterior parameters of the second layer process and the means and variances of the distributions. The parameters of the first postulated layer are initialized to random values, while for the hyperparameters we use \textit{ad-hoc} uninformative values, similar to \cite{studentshmm}. Specifically, the hyperparameters of the Dirichlet priors are set to very small values ($10^{-3}$). For the parameters of the emission distributions, we set $\lambda_{ij} = \tilde{\lambda}_{ij}=0.25$ and $\eta_{ij} = \tilde{\eta}_{ij}= D + 2$, where $D$ is the dimensionality of the data.
	%
	
	\subsection{States, Mixtures and Temporal Dependencies}
	We trained different models with varying number of states and mixtures to investigate their effect on model effectiveness. We observed that the recognition accuracy is not considerably sensitive to the chosen configuration; thus, we run multiple configurations and retain the one with the higher ELBO. For all experiments, we set $K=2$, as we observed that the inferred temporal dependence posteriors rarely assigned high probability (if any) to $p(z_t>2|z_{t-1})$; this behavior is reasonable considering the physical meaning of the used data features. Nevertheless, we did not observe any drop in accuracy when setting $K>2$. For $K=1$, the model reduces to a simple HMM, and exhibits a significant drop in recognition accuracy. This outcome strongly corroborates the benefits of our dynamic temporal dependency inference approach. 
	\subsection{Experimental Results}
	We follow the train and test splits as suggested by the authors of the respective datasets. We train one model per class and assign the test data to the class with the maximum predictive posterior (Eq. \eqref{eqn:predictive}). We compare our approach to related approaches, along with a comparison with HDM \cite{hdm}. Out of the HDM model variants therein, we focus on the BV variant for comparability, whereby variational inference is utilized for computing the predictive likelihood.
	\begin{table}
		\caption{Recognition Accuracy ($\%$) for individual dataset experiments.}
		\label{experiments:individual}
		\begin{center}
			\renewcommand*{\arraystretch}{1.1}
			\resizebox{0.5\linewidth}{!}{%
				\begin{tabular}{|c|c|c|c|c|c|}
					\hline
					Model & MSRA & UTD & G3D & Penn & Avg. \\\hline
					HMM & 67.8 & 82.8 & 68.1 & 82.3 & 75.3 \\\hline
					HMM\textsuperscript{2} & 80.2 & 83.1 & 82.6 & 84.4 & 82.6\\\hline
					HSMM & 66.3 & 82.3 & 77.5 & 78.9 & 76.35 \\\hline
					LSTM & 74.7 & 77.0 & 82.2 & 90.3 & 81.1 \\\hline
					HDM-PI & 70.3 & 84.4 & 79.4 & 89.8 & 81.0 \\\hline
					HDM-PL & 80.6 & 90.2 & 87.7 & 91.6 & 87.5 \\\hline
					HDM-BV & 82.1 & 91.4 & 87.7 & 90.8 & 88.0 \\\hline
					VB-CD-HMM & $\mathbf{82.5}$ & $\mathbf{92.7}$ & $\mathbf{90.6}$ & $\mathbf{92.0}$ & $\mathbf{89.45}$ \\\hline
				\end{tabular}
			}
		\end{center}
	\end{table}
	%
	%
	\\[1em]
	\noindent \textbf{Individual Datasets.}
	We first consider the MRSA dataset, where the recognition rates of our approach can be found in the first column of Table \ref{experiments:individual}. Compared to baseline models such as HMMs, HSMMs and LSTMs, our model performs much better in classification accuracy with an average improvement of $12.9\%$. Even though the recently proposed HDM \cite{hdm} consistently improves over the considered alternatives, it falls short compared to the proposed VB-CD-HMM model. This behavior is consistent across all the considered datasets, resulting in an average classification accuracy of $89.45\%$, outperforming HDM by $1.45\%$. Additionally, we observe that our proposed approach consistently and significantly outperforms a second-order HMM (HMM\textsuperscript{2}) evaluated under the same experimental and modeling setup. Even though HMM\textsuperscript{2} improves over simple first-order HMMs, by considering more complex dynamics, it still falls short to the flexibility of modeling dependencies over long horizons, that VB-CD-HMM offers. Note also that the prowness of HMM\textsuperscript{2} comes at the cost of a considerable increase in the resulting computational complexity; this is in contrast to our model, which still relies on (a variant of) the forward-backward algorithm.

	We additionally compare the recognition accuracy of our approach to other state-of-the-art methods. The corresponding performances are illustrated in Table \ref{experiments:stateoftheart}. As we observe, our approach yields clearly improved accuracy compared to the competition for UTD, Penn and NTU datasets. For the G3D dataset, our method slightly outperforms LRBM \cite{lrbm}, but R3DG \cite{r3dg} performs better; this is due to the sophisticated feature engineering and combination of several approaches in contrast to our simple joints locations and motion features.

	\begin{table}
		\caption{Recognition accuracy for all the considered datasets compared to state-of-the-art.}
		\label{experiments:stateoftheart}
		\renewcommand*{\arraystretch}{1.2}
		\resizebox{1.\linewidth}{!}{%
			\begin{tabular}{c|c|c|c|c|c}
				\hline
				\multicolumn{6}{c}{Dataset}\\\hline
				\multicolumn{6}{c}{Method $\parallel$ Acc}\\\hline
				MSRA & UTD & G3D & Penn & NTU (x-view) & NTU (x-subject)\\\hline
				ST-LSTM\cite{stlstm} $\parallel$ \textbf{94.8} & Fusion\cite{UTD} $\parallel$ 79.1 & LRBM \cite{lrbm} $\parallel$ 90.5 & Actemes\cite{Penn} $\parallel$ 86.5  & ST-TSL \cite{sttsl} $\parallel$ 92.4 & ST-TSL\cite{sttsl} $\parallel$ 84.8\\
				B-GC-LSTM\cite{zhao_iccv} $\parallel$ 94.5 & SOS-CNN\cite{soscnn} $\parallel$ 87.0 & R3DG\cite{r3dg} $\parallel$ \textbf{91.1} & AOG\cite{AOG} $\parallel$ 84.8 & B-GC-LSTM \cite{zhao_iccv} $\parallel$ 89.0 & B-GC-LSTM \cite{zhao_iccv} $\parallel$ 81.8\\
				VB-CDHMM $\parallel$ 92.5 & VB-CDHMM $\parallel$ \textbf{92.7} & VB-CDHMM $\parallel$ 90.6 & VB-CDHMM $\parallel$ \textbf{92.0} & VB-CDHMM $\parallel$ \textbf{93.0} & VB-CDHMM $\parallel$ \textbf{85.0}\\\hline

			\end{tabular}
		}
	\end{table}
	The main focus of this Section is to compare and prove the efficacy of our approach within the family of more general HMM-based methods in the tackled application domain. Thus, we did not explore complicated model-based and specific-task tailored feature engineering; this allows for both transparency of comparison to recent related published work, and interpretability of our results. Nevertheless, as we shall see in the next section, our method clearly outperforms R3DG, when we randomly omit observations; this behavior vouches for the improved robustness of our approach. Likewise, ST-LSTM \cite{stlstm} incorporates complicated feature extraction mechanisms, thus explaining the performance gap compared to our approach. For the NTU dataset \cite{ntu}, to the best of our knowledge, there are no results for recent HMM-based methods. As we observe, compared to the state-of-the-art methods ST-LSTM \cite{stlstm} and B-GC-LSTM \cite{zhao_iccv}, VB-CD-HMM yields considerable improvements in both splits, despite its simplicity and lower computational complexity.

	\subsection{Missing Values}
	Generative models come with the additional benefit of robustness to missing values. This advantage is of great importance in the human action recognition field, especially when using skeletal data, where the dataset may be corrupted with missing observations, e.g., due to occlusion. As an HMM variant, the proposed model is such a generative model; we thus assess the efficacy of our approach when missing values are present. To this end, we train the considered architecture in the UTD, MSRA and G3D datasets, with randomly missing values from both the train and test data. Analogous experiments have been performed in \cite{hdm}, and we adopt the same procedure, including the proportion of missing values with respect to the original data. 

	\begin{table*}
		
		\caption{Recognition Accuracy ($\%$) with missing values.}
		\label{experiments:missing}
		\begin{center}
			\resizebox{.7\linewidth}{!}{
				\begin{tabular}{|l|l|ccc|ccc|ccc|}
					\hline
					\multicolumn{2}{|c|}{Dataset} & \multicolumn{3}{c|}{UTD} & \multicolumn{3}{c|}{MSRA} & \multicolumn{3}{c|}{G3D}\\\hline
					\multicolumn{2}{|c|}{Missing Portion} & $10\%$    & $30\%$ & $50\%$ & $10\%$    & $30\%$ & $50\%$ & $10\%$    & $30\%$ & $50\%$   \\
					\hline
					\parbox[t]{2mm}{\multirow{4}{*}{\rotatebox[origin=c]{90}{Model}}} & R3DG \cite{r3dg} & $81.5$ & $74.0$ & $72.0$ 
					& $78.0$ & $72.0$ & $70.0$
					& $87.0$ & $86.0$ & $83.0$\\
					& DLSTM \cite{dlstm} & $70.5$ & $66.0$ & $63.0$ 
					& $68.0$ & $63.0$ & $61.0$
					& $81.0$ & $76.0$ & $73.0$\\
					& HDM \cite{hdm} & $91.0$ & $90.5$ & $90.0$ 
					& $80.5$ & $78.0$ & $76.0$
					& $90.0$ & $89.0$ & $88.0$\\
					& VB-CD-HMM & $\mathbf{92.55}$ & $\mathbf{91.6}$ & $\mathbf{90.2}$ 
					& $\mathbf{81.7}$ & $\mathbf{80.1}$ & $\mathbf{79.1}$
					& $\mathbf{90.2}$ & $\mathbf{89.3}$ & $\mathbf{88.5}$\\
					\hline
				\end{tabular}
			}
		\end{center}
	\end{table*}

\begin{figure*}
		\centering
 		\includegraphics[width=0.7\linewidth]{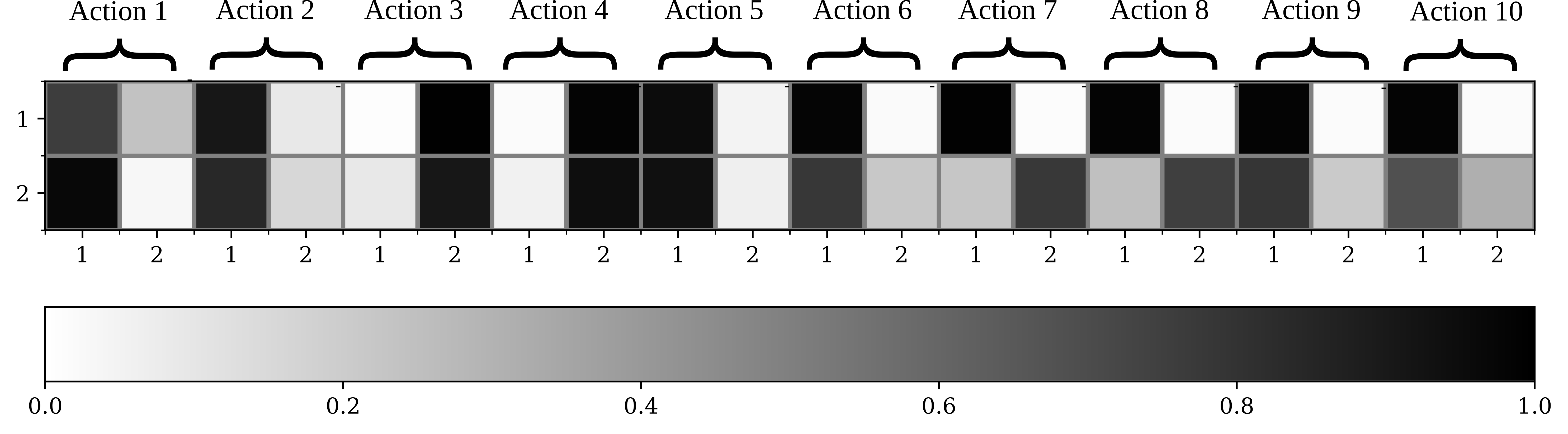}
		\caption{The inferred temporal dependence probabilities $A_{kk'}$ for the first $10$ actions in the UTD dataset. Black denotes very high probability, while white very low. Actions with similar patterns, e.g., 1 (swipe left), 2 (swipe right), 5 (throw), i.e., comprising sequential movements to “one” direction, are dominantly dependent on the previous frame; in contrast, more complicated movements, e.g., 3 (wave), 4 (clap), mainly depend on two steps back, possibly dominated by the reversal in direction. Even more complicated actions, as 7 (basketball shoot) and 8 (draw x) exhibit more complex temporal dependencies, that are captured via the resulting inferred patterns.}
		\label{plot:dependence}
	\end{figure*}

	Thus, in Table \ref{experiments:missing}, we report the recognition rates when we randomly omit $10\%, 30\%$ and $50\%$ of the observations. As we observe, our method clearly outperforms the R3DG \cite{r3dg} and DLSTM \cite{dlstm} methods by a clear margin. The same behavior is observed compared to the more relative HDM approach \cite{hdm}. Note that even though the reported HDM accuracies are obtained through Gibbs sampling (BG variant), our approach exhibits the smallest decrease in recognition accuracy relative to the increase of missing values. 
	%
	%
	\\\\
	\noindent \textbf{Further Insights.}
	Finally, we examine the patterns of the first layer process, that is the \textit{dependence generator} layer of the VB-CD-HMM model. We gain further insights to the behavior of the model and assert that the temporal dependencies do not collapse to simple first order dynamics (which would effectively reduce the model to a conventional HMM). To this end, we focus on a trained model on the \textit{UTD} dataset and choose a subset of ten actions to investigate their posterior parameters concerning the generation of dependencies. As is clearly shown in Fig. \ref{plot:dependence}, the distribution of temporal dependencies is essentially different for each action, providing strong empirical evidence that the introduced mechanism can capture complex varying temporal dynamics in the data. 
	\section{Conclusions}
	Our experimental results have provided strong empirical evidence for the efficacy of our approach, yielding competitive recognition accuracy in the Human Action Recognition task. Compared to the commonly considered LSTM-based implementations, our proposed architecture:
	\begin{itemize}[noitemsep,topsep=0pt, itemsep=0pt, topsep=0pt, parsep=0pt]
		\item offers the flexibility to model complex sequential data accompanied by the interpretability of its discrete bottlenecks; this is in stark contrast to LSTM-based methods that assume continuous states that essentially entail no probabilistic interpretation.
		\item enjoys closed-form updates that allow for significantly lower computational and training complexity, devoid of complications, e.g., vanishing gradients, and
		\item can effectively handle data with missing values, exhibiting the \textit{smallest decrease} in recognition performance relative to other approaches.  
	\end{itemize}
	We emphasize that the proposed mechanism for inferring the dependencies between the states in our proposed model allows for analyzing the inferred temporal dynamics. Thus, we can gain significant insights and examine how the dynamics adapt to the complexity and variations of different train/test sequences. Therefore, this capacity can be of great benefit to the current effort of the community towards interpretable machine learning models.

%
%
%
\bibliographystyle{splncs04}
\bibliography{cdhmm}

\end{document}

%% file: graphs/cdhmm_graph.tex
\begin{tikzpicture}
  \node[main] (z1) [minimum size=1.cm]{$z_{t-1}$};
  \node[main] (z2) [right=of z1, minimum size=1.cm] {$z_{t}$};
  \node[main] (z3) [right=of z2, minimum size=1.cm] {$z_{t+1}$};
  
  \node[main] (x1) [below= of z1, minimum size=1.cm] {$x_{t-1}$};
  \node[main] (x2) [right=of x1, minimum size=1.cm] {$x_{t}$};
  \node[main] (x3) [right=of x2,minimum size=1.cm] {$x_{t+1}$};
  
  \node[right= of x3] (dotsx_right) {$\dots$};
  \node[left= of x1] (dotsx_left) {$\dots$};
  
  \node[main] (x4) [left=of dotsx_left, minimum size=1.cm] {$x_{\scaleto{t-K+1}{3.5pt}}$};
  \node[main] (x5) [left=of x4, minimum size=1.cm] {$x_{t-K}$};
 
   \node[main,fill=black!2] (y1) [below=of x1,minimum size=1.cm] {$y_{t-1}$};
  \node[main,fill=black!2] (y2) [below=of x2,minimum size=1.cm] {$y_t$};
  \node[main,fill=black!2] (y3) [below=of x3, minimum size=1.cm] {$y_{t+1}$};
  \node[main,fill=black!2] (y4) [below=of x4, minimum size=1.cm] {$y_{\scaleto{t-K+1}{3.5pt}}$};
  \node[main,fill=black!2] (y5) [below=of x5, minimum size=1.cm] {$y_{t-K}$};
  
  \node[right= of y4] (dotsy_right) {$\dots$};

  \path (z1) edge [connect] (z2)
        (z2) edge [connect] (z3);
        
  \path (z1) edge [connect] (x1)
        (z2) edge [connect] (x2)
        (z3) edge [connect] (x3);
  
  \path (x1) edge [connect] (x2)
        (x2) edge [connect] (x3)
        (x1) edge [connect, bend left] (x3)
        (x4) edge [connect, bend left] (x3)
        
        (x4) edge [connect, bend left] (x2)
        (x5) edge [connect, bend left] (x2)
        (x4) edge [connect, bend left] (x1)
        (x5) edge [connect, bend left] (x1);

  \path (x1) edge [connect] (y1);
  \path (x2) edge [connect] (y2);
  \path (x3) edge [connect] (y3);
  \path (x4) edge [connect] (y4);
  \path (x5) edge [connect] (y5);

\end{tikzpicture}

%% file: graphs/vb_cdhmm.tex
\begin{tikzpicture}

  
  \node[main] (z1) [minimum size=1.cm]{$z_{t-1}$};
  \node[main] (z2) [right=of z1, minimum size=1.cm] {$z_{t}$};
  \node[main] (z3) [right=of z2, minimum size=1.cm] {$z_{t+1}$};
  
  \node[left= of z1] (dotsz_left) {$\dots$};
    
  \node[main] (z4) [left= of dotsz_left, minimum size=1.cm] {$z_{\scaleto{t-K+1}{3.5pt}}$}; 
  \node[main] (z5) [left= of z4, minimum size=1.cm] {$z_{t-K}$};
  \node (dotsz_left_left) [left= of z5, minimum size=1.cm] {$\dots$};
  \node[main] (z6) [left= of dotsz_left_left, minimum size=1.cm] {$z_1$};  
  
  \node[main] (varpi_hat) [left= of z6] {$\hat{\pi}$};
  \node[box, rounded corners, draw, left=0.75cm of z6, minimum size = 1.2cm] (hat_varpi_plate) {};
  
  \node[main] (pi_hat) [above= of varpi_hat] {$\hat{A}$};
  \node[box, rounded corners, above=0.75cm of varpi_hat, draw, minimum size = 1.2cm] (hidden_plate) {};

  \node (alpha0) [left=of varpi_hat] {$\alpha_0$};
  \node(alpha) [left= of pi_hat] {$\alpha$};

  \node[main] (x1) [below= 2.5cm of z1, minimum size=1.cm] {$x_{t-1}$};
  \node[main] (x2) [right=of x1, minimum size=1.cm] {$x_{t}$};
  \node[main] (x3) [right=of x2,minimum size=1.cm] {$x_{t+1}$};
  
  \node[right= of x3] (dotsx_right) {$\dots$};
  \node[left= of x1] (dotsx_left) {$\dots$};
    
  \node[main] (x4) [left=of dotsx_left, minimum size=1.cm] {$x_{\scaleto{t-K+1}{3.5pt}}$};
  \node[main] (x5) [left=of x4, minimum size=1.cm] {$x_{t-K}$};
  \node (dotsx_left_left) [left= of x5, minimum size=1.cm] {$\dots$};
  \node[main] (x6) [left= of dotsx_left_left, minimum size=1.cm] {$x_1$};

  \node[main] (varpi) [left= of x6] {$\pi$};
  \node[box, rounded corners, draw, left=0.75cm of x6, minimum size = 1.2cm] (varpi_plate) {};

  \node[main] (pi) [above= 0.8cm of varpi] {$A$};
  \node[box, rounded corners, draw, above=0.55cm of varpi, minimum size = 1.2cm] (pi_plate) {};
  
  \node (eta0) [left=of varpi] {$\eta_0$};
  \node(eta) [left= of pi] {$\eta$};

  \node[main,fill=black!2] (y1) [below=2.5cm of x1,minimum size=1.cm] {$y_{t-1}$};
  \node[main,fill=black!2] (y2) [right=of y1,minimum size=1.cm] {$y_t$};
  \node[main,fill=black!2] (y3) [right=of y2, minimum size=1.cm] {$y_{t+1}$};
  \node[left= of y1] (dotsy_left) {$\dots$};
    
  \node[main,fill=black!2] (y4) [left= of dotsy_left, minimum size=1.cm] {$y_{\scaleto{t-K+1}{3.5pt}}$}; 
  \node[main,fill=black!2] (y5) [left= of y4, minimum size=1.cm] {$y_{t-K}$};
  \node (dotsy_left_left) [left= of y5, minimum size=1.cm] {$\dots$};
  \node[main,fill=black!2] (y6) [left= of dotsy_left_left, minimum size=1.cm] {$y_1$};  
  
  \node[main] (theta) [above left= 0.3cm and 1.3cm of y6] {$\scaleto{\phi_{im}}{7.5pt}$};
  \node[box, rounded corners, draw, above left=0cm and 1.cm of y6, minimum size = 1.2cm] (obs_plate) {};
  \node (prior_theta) [left= of theta] {$\lambda, m, \eta, S$};
 
  
  \path (alpha) edge [connect] (pi_hat)
        (alpha0) edge [connect] (varpi_hat)
        (varpi_hat) edge [connect] (z6)
        (pi_hat.east) edge [connect, bend left=15] (z5)
        (pi_hat.east) edge [connect, bend left=15] (z4)
        (pi_hat.east) edge [connect, bend left=15] (z3)
        (pi_hat.east) edge [connect, bend left=15] (z2)
        (pi_hat.east) edge [connect, bend left=15] (z1);

  \path (eta) edge [connect] (pi)
        (eta0) edge [connect] (varpi)
        (varpi) edge [connect] (x6)
        (pi.east) edge [connect, bend left=20] (x5.north)
        (pi.east) edge [connect, bend left=20] (x4.north)
        (pi.east) edge [connect, bend left=20] (x3.north)
        (pi.east) edge [connect, bend left=20] (x2.north)
        (pi.east) edge [connect, bend left=20] (x1.north);

  \path (prior_theta) edge [connect] (theta)
        (theta) edge [connect, bend left=20] (y6.north)
        (theta) edge [connect, bend left=20] (y5.north)
        (theta) edge [connect, bend left=20] (y4.north)
        (theta) edge [connect, bend left=20] (y3.north)
        (theta) edge [connect, bend left=20] (y2.north)
        (theta) edge [connect, bend left=20] (y1.north);
  
  \path (z1) edge [connect] (z2)
        (z2) edge [connect] (z3)
        (z6) edge [connect] (dotsz_left_left)
        (dotsz_left_left) edge [connect] (z5)
        (z5) edge [connect] (z4)
        (z4) edge [connect] (dotsz_left)
        (dotsz_left) edge [connect] (z1);
        
  \path (z1) edge [connect] (x1)
        (z2) edge [connect] (x2)
        (z3) edge [connect] (x3)
        (z4) edge [connect] (x4)
        (z5) edge [connect] (x5)
        (z6) edge [connect] (x6);
  
  \path (x1) edge [connect] (x2)
        (x2) edge [connect] (x3)
        (x2) edge [->, bend left] (x3)
        (x1) edge [->, bend left] (x3)
        (x4) edge [->, bend left] (x3)
        (x4) edge [->, bend left] (x2)
        (x5) edge [->, bend left] (x2)
        (x4) edge [->, bend left] (x1)
        (x5) edge [->, bend left] (x1)
        (x6) edge [connect] (dotsx_left_left)
        (dotsx_left_left) edge [connect] (x5)
        (x5) edge [connect] (x4)
        (x4) edge [connect] (dotsx_left)
        (dotsx_left) edge [connect] (x1);

  \path (x1) edge [connect] (y1);
  \path (x2) edge [connect] (y2);
  \path (x3) edge [connect] (y3);
  \path (x4) edge [connect] (y4);
  \path (x5) edge [connect] (y5);
  \path (x6) edge [connect] (y6);

\end{tikzpicture}

%% file: main.bbl
\begin{thebibliography}{10}
\providecommand{\url}[1]{\texttt{#1}}
\providecommand{\urlprefix}{URL }
\providecommand{\doi}[1]{https://doi.org/#1}

\bibitem{review3d}
Aggarwal, J., Xia, L.: Human activity recognition from 3d data: A review.
  Pattern Recognition Letters  \textbf{48} (10 2014)

\bibitem{mariRobotics}
Aycard, O., Mari, J., Washington, R.: Learning to automatically detect features
  for mobile robots using second-order hidden markov models. IJARS  (2005)

\bibitem{Begleiter}
Begleiter, R., El-Yaniv, R., Yona, G.: On prediction using variable order
  markov models. J. Artif. Int. Res.  \textbf{22} (Dec 2004)

\bibitem{g3d}
{Bloom}, V., {Makris}, D., {Argyriou}, V.: G3d: A gaming action dataset and
  real time action recognition evaluation framework. In: CVPRW (June 2012)

\bibitem{buhlmann1999}
Bühlmann, P., Wyner, A.J.: Variable length markov chains. Ann. Statist.
  (1999)

\bibitem{chatzis_gaussian}
{Chatzis}, S.P., {Demiris}, Y.: Nonparametric mixtures of gaussian processes
  with power-law behavior. IEEE Transactions on Neural Networks and Learning
  Systems  \textbf{23}(12),  1862--1871 (2012)

\bibitem{gplvm}
{Chatzis}, S.P., {Kosmopoulos}, D.: A latent manifold markovian dynamics
  gaussian process. IEEE Transactions on Neural Networks and Learning Systems
  (2015)

\bibitem{chatzisNonstationary}
Chatzis, S., Kosmopoulos, D., Papadourakis, G.: A nonstationary hidden markov
  model with approximately infinitely-long time-dependencies. IJAIT
  \textbf{25} (2016)

\bibitem{studentshmm}
Chatzis, S.P.: A variational bayesian methodology for hidden markov models
  utilizing student's-t mixtures. Pattern Recognition  \textbf{44} (2011)

\bibitem{chatzisInfinitely}
Chatzis, S.P.: Margin-maximizing classification of sequential data with
  infinitely-long temporal dependencies. Expert Systems with Applications
  \textbf{40} (2013)

\bibitem{UTD}
{Chen}, C., {Jafari}, R., {Kehtarnavaz}, N.: Utd-mhad: A multimodal dataset for
  human action recognition utilizing a depth camera and a wearable inertial
  sensor. In: Procs. ICIP (Sep 2015)

\bibitem{dimitrakakis10a}
Dimitrakakis, C.: Bayesian variable order markov models. In: Procs. ICAIS 13

\bibitem{ENGELBRECHT}
Engelbrecht, H., du~Preez, J.: Efficient backward decoding of high-order hidden
  markov models. Pattern Recognition  \textbf{43} (2010)

\bibitem{soscnn}
{Hou}, Y., {Li}, Z., {Wang}, P., {Li}, W.: Skeleton optical spectra-based
  action recognition using convolutional neural networks. IEEE TCSVT
  \textbf{28}(3),  807--811 (2018)

\bibitem{msra}
Li, W., Zhang, Z., Liu, Z.: Action recognition based on a bag of 3d points. In:
  IEEE CVPRW 2010 (07 2010)

\bibitem{stlstm}
Liu, J., Shahroudy, A., Xu, D., Kot, A., Wang, G.: Skeleton-based action
  recognition using spatio-temporal lstm network with trust gates. IEEE
  Transactions on Pattern Analysis and Machine Intelligence  \textbf{40},
  3007--3021 (12 2018)

\bibitem{Nel}
{Nel}, E., {du Preez}, J.A., {Herbst}, B.M.: Estimating the pen trajectories of
  static signatures using hidden markov models. IEEE Transactions on Pattern
  Analysis and Machine Intelligence  \textbf{27} (Nov 2005)

\bibitem{AOG}
{Nie}, B.X., {Xiong}, C., {Zhu}, S.: Joint action recognition and pose
  estimation from video. In: Procs. CVPR. pp. 1293--1301 (June 2015)

\bibitem{lrbm}
Nie, S., Wang, Z., Ji, Q.: A generative restricted boltzmann machine based
  method for high-dimensional motion data modeling. Comput. Vis. Image Underst.
   (2015)

\bibitem{panousis}
Panousis, K., Chatzis, S., Theodoridis, S.: Nonparametric {B}ayesian deep
  networks with local competition. Procs ICML, PMLR (June 2019)

\bibitem{ntu}
{Shahroudy}, A., {Liu}, J., {Ng}, T., {Wang}, G.: Ntu rgb+d: A large scale
  dataset for 3d human activity analysis. In: Procs. CVPR. pp. 1010--1019
  (2016)

\bibitem{sttsl}
Si, C., Jing, Y., Wang, W., Wang, L., Tan, T.: Skeleton-Based Action
  Recognition with Spatial Reasoning and Temporal Stack Learning: 15th EC,
  Munich, Germany, September, 2018, Proceedings, Part I, pp. 106--121 (09 2018)

\bibitem{theodoridis}
Theodoridis, S.: Machine Learning: A Bayesian and Optimization Perspective.
  Academic Press, Inc., USA, 2nd edn. (2020)

\bibitem{r3dg}
{Vemulapalli}, R., {Arrate}, F., {Chellappa}, R.: Human action recognition by
  representing 3d skeletons as points in a lie group. In: Procs. CVPR (June
  2014)

\bibitem{surveyAction}
Weinland, D., Ronfard, R., Boyer, E.: A survey of vision-based methods for
  action representation, segmentation and recognition. Comput. Vis. Image
  Underst.  (2011)

\bibitem{hsmm}
Yu, S.Z.: Hidden semi-markov models. Artif. Intell.  \textbf{174} (2010)

\bibitem{Penn}
Zhang, W., Zhu, M., Derpanis, K.G.: From actemes to action: A
  strongly-supervised representation for detailed action understanding. In:
  ICCV (Dec 2013)

\bibitem{zhao_iccv}
{Zhao}, R., {Wang}, K., {Su}, H., {Ji}, Q.: Bayesian graph convolution lstm for
  skeleton based action recognition. In: Procs. ICCV. pp. 6881--6891 (2019)

\bibitem{hdm}
{Zhao}, R., {Xu}, W., {Su}, H., {Ji}, Q.: Bayesian hierarchical dynamic model
  for human action recognition. In: Procs. CVPR (June 2019)

\bibitem{dlstm}
Zhu, W., Lan, C., Xing, J., Zeng, W., Li, Y., Shen, L., Xie, X.: Co-occurrence
  feature learning for skeleton based action recognition using regularized deep
  lstm networks. In: Procs. AAAI (2016)

\end{thebibliography}
